\documentclass{article}

\usepackage[nonatbib, final]{nips_2018}

\usepackage[utf8]{inputenc} 
\usepackage[T1]{fontenc}    
\usepackage{hyperref}       
\usepackage{url}            
\usepackage{booktabs}       
\usepackage{amsfonts}       
\usepackage{nicefrac}       
\usepackage{microtype}      
\usepackage{amsmath}
\usepackage{graphicx}

\usepackage{hyperref}
\usepackage{cleveref}
\usepackage[square,sort,comma,numbers]{natbib}

\title{A Deep 2-Dimensional Dynamical Spiking Neuronal Network for Temporal Encoding trained with STDP}

\author{
  Matthew S. Evanusa\\
  Dept. of Computer Science\\
  University of Maryland\\
  \texttt{evanusa@cs.umd.edu} \\
  \And
  Yiannis Aloimonos\\
  Dept. of Computer Science\\
  University of Maryland\\
  \texttt {yiannis@cs.umd.edu}\\
  \And
  Cornelia Fermuller\\
  Dept. of Computer Science\\
  University of Maryland\\
  \texttt {fer@umiacs.umd.edu}
}
\usepackage{natbib}
\usepackage{cleveref}

\begin{document}

\maketitle

\begin{abstract}
The brain is known to be a highly complex, asynchronous dynamical system that is highly tailored to encode temporal information. However, recent deep learning approaches to not take advantage of this temporal coding. Spiking Neural Networks (SNNs) can be trained using biologically-realistic learning mechanisms, and can have neuronal activation rules that are biologically relevant. This type of network is also structured fundamentally around accepting temporal information through a time-decaying voltage update, a kind of input that current rate-encoding networks have difficulty with.  Here we show that a large, deep layered SNN with dynamical, chaotic activity mimicking the mammalian cortex with biologically-inspired learning rules, such as STDP, is capable of encoding information from temporal data.  We argue that the randomness inherent in the network weights allow the neurons to form groups that encode the temporal data being inputted after self-organizing with STDP.  We aim to show that precise timing of input stimulus is critical in forming synchronous neural groups in a layered network. We analyze the network in terms of network entropy as a metric of information transfer.  We hope to tackle two problems at once: the creation of artificial temporal neural systems for artificial intelligence, as well as solving coding mechanisms in the brain.

\end{abstract}

\section{Background and Introduction}

Many of the major advances in A.I. have been made due to reverse engineering biological brains, including Deep Neural Networks (DNNs) and Reinforcement Learning (RL) (\cite{mnih2015human}). Research into A.I. has been given a boost in recent decades due to success in reverse engineering neural connections and functions of the mammalian cortex, and in reverse, machine learning approaches have helped computational neuroscientists in learning spike patterns from neural data.  So-called “deep learning” frameworks (\cite{krizhevsky2012imagenet}) adopt the feed-forward, layered aspects of the connections in the brain, while some (convolutional networks) go further to mimic the convolutional connections that are present in many pathways.  These approaches have demonstrated human level accuracy in some tasks, such as face detection from static images (\cite{taigman2014deepface}).  However, because they rely on non-biological learning mechanisms, it is difficult to use these networks to uncover biological mechanisms for neural information encoding.  In addition, these networks still struggle to classify temporal data; this will be a critical next step for A.I. systems of the future. In the neuroscience community, it is a current debate as to whether the information is encoded in the firing rate of the neuron, or if the information is encoded in the precise timing of incoming spikes (\cite{mainen1995reliability,van1997reproducibility, vanrullen2005spike}).  It has been shown that unsupervised learning rules, such as STDP (section 1.4), are capable of encoding this spike-timing information into the synaptic weights (\cite{gerstner1997neural,izhikevich2004spike}).  It has also been argued that through these learning rules, the neurons compete with one another to form groups in a selection process (\cite{edelman1987neural}). Here we investigate whether biologically-inspired neural plasticity rules combine with precise timing of incoming firing rates to encode information about novel stimuli in the form of a biologically-inspired camera in a multi-layered network.  

\subsection{Rate-Encoding Models}

Current ANNs, including deep-learning frameworks, are what computational neuroscientists refer to as "rate-encoding models", in that the real-value of the output from a unit (roughly approximating an integrating neuron) represents a firing rate, or a probability of firing, over time. These networks accept inputs all at once and are also referred to as "linear non-linear" models (\cite{ostojic2011spiking}), in that they represent a combination of a linear operation (summation of incoming connections), as well as a non-linear operation called the "activation function" (some sort of sigmoidal, inverse tangent function, or rectified linear unit).  The rate $R$ of a neuron's fire is algebraically:

\begin{equation}
R= N / T
\end{equation}

for N spikes in a given time-scale window T. As we can see from this interpretation, these rate-encoding networks \textit{divide out} time as a factor in their design; they are fundamentally time-less.  Any rate-encoding network that attempts to encode a sequence of images as a video is not encoding 2-dimensional images in time, but rather the entire sequence of 2-dimensional frames at once as one 3-dimensional feature block.   

\subsection{Dynamical and Reservoir Computing}

Our proposed network is closer to what is referred to as "reservoir computing" (\cite{lukovsevivcius2009reservoir}), in that it acts as a chaotic, dynamic system due to the neuron update, lateral connections, and randomness placed in the network.  Reservoir computers are a type of recurrent neural network that do not learn on the recurrent connections within the random recurrent connections but rather only on a read-out layer. Reservoir computers have demonstrated high success in classification of difficult temporal patterns due to the integration of time information in each artificial neuron's update rule.  They have been shown to encode very complicated temporal input, even input that exhibits very chaotic patterns such as the Lorentz attractor.  Although reservoir computers require a strong supervised signal to drive them towards a target, they still demonstrate that random weights and connections are a powerful factor in a network's ability to encode temporal data. As such, they have been successfully used to analyze complicated recurrent neuronal dynamics in decision making (\cite{chaisangmongkon2017computing,enel2016reservoir}). 

\subsection{Spiking Neural Networks}

The so-called “third generation” of neural networks, SNNs offer a promising way of encoding time information, although the dynamics are hardly new (\cite{hodgkin1952currents}), and what constitutes a "spiking neural network" is broad, as well as the desired goal of biological plausibility or classifying power.  SNNs remain at this point a loose collection of different neural-like networks that share the common feature that individual neurons operate as dynamical systems that individually collect voltage and “spike” an output at a given threshold, but differ in their layout and learning rules.  SNN activation functions for the neuron vary widely, from neurons that simply integrate over time without any voltage leak (\cite{urbanczik2001similar}) to those that model ion channels at individual compartments along the neuron's dendrite (\cite{segev1989compartmental}).  Analysis of the interconnection of these spiking units in larger networks, however, is a more recent undertaking (\cite{ghosh2009spiking}).  Efforts have been made to show that backpropagation can be implemented with SNNs as well (\cite{bohte2002error}), although this departs from the biological plausibility and principles of self-organization.  Here we analyze the information-encoding potential of a semi-recurrent feed-forward cortical-like deep network of spiking neurons, and investigate whether information is stored in assemblies as Hebb postulated. We take the approach of using a network that has solid biological basis while trying to keep the operations as simple as possible.

\subsection{STDP and iSTDP}
Donald Hebb postulated that given a neuron’s spiking behavior, which is exhibited in SNNs, neurons organize into groups (what he called cell assemblies) that encode information (\cite{hebb2005organization}).  Hebb formulated a learning rule that has been boiled down to the phrase "neurons that fire together, wire together.", and is formalized for continuous (rate-encoding) networks as:

\begin{equation}
\delta w_{1,2} = \alpha N_1 N_2
\end{equation}

For neurons $N_1$ and $N_2$, with learning rate $\alpha$; the weight $w$ changes only when both neurons fire.  This rule allows a networks' neurons, either rate-encoding or spiking, to learn in an unsupervised manner to reorganize their connections to match the input as there is no error term.  However, this simple rule has been extensively studied and proven to be able to self-classify input data, for example in Hopfield Networks, although they have issues such as a limited patterns storage capacity(\cite{abu1985information}). Recent discoveries showed that neurons in the brain modify their connections via a \textit{temporally asymmetric} learning rule, dubbed Spike Time Dependent Plasticity (\cite{bi2001synaptic,sejnowski1999book}), a timing-based modification of Hebb's original learning rule that takes advantage of (and requires) spiking dynamics. STDP combines two phenomena that occur in neurons in vivo: Long Term Potentiation (LTP), which strengthens the connection (weight) between two neurons, and Long Term Depression (LTD), which weakens it.  The rule is implemented here using a common trace mechanism for STDP:

\begin{equation}
	\delta w_{pre,post} = 
	\begin{cases} 
      A_+w_{pre,post}T_{post} & \text{if $N_{pre}$ fires} \\
      A_-w_{pre,post}T_{pre} & \text{if $N_{post}$ fires} \\
   \end{cases}
\end{equation}

for presynaptic neuron $N_{pre}$ and post $N_{post}$. $T$ is a trace that goes to some maximum value (this paper uses 2) and exponentially decays after the neuron fires, with time constant $\tau$. $A_+$ and $A_-$ are scaling factors that allow for the maximum LTP and LTD, respectively, and can be weighted.  The maximum excitatory and inhibitory weights were bounded; this is a natural balancing mechanism inspired by the fact that a given neuron can only put a maximum number of transmitter receptors physically on the outside of the neuron, and from studies that show that LTP works slower in high firing regimes \cite{watt2010homeostatic}.  STDP increases the weights for a connection between two neurons, with a given pre- and post-synaptic connection, only when the presynaptic neuron fires followed closely in time by a firing of the postsynaptic neuron. If a postsynaptic neuron fires before a presynaptic neuron, that connection is weakened (\cite{bi2001synaptic}). Because some synapses are also weakened, this results in a competitive process whereby only the connections that fire in lock-step with the input are strengthened.  It has also been shown that STDP could be the fundamental mechanism that causes neurons, both in the brain and SNNs, to organize into Hebbian assemblies (\cite{izhikevich2004spike}). It has already been shown that STDP is a powerful enough learning rule to allow network to categorize temporal data in the form of EEG (\cite{kasabov2012evolving}), although the network setup was quite different from what will be described in Methods.  Previous work also showed a 2-dimensional layered network that was able to classify static images of handwritten digits using STDP on a more simplified network (\cite{iakymchuk2015simplified}).

However, it has also been argued that STDP alone is not enough to maintain a balanced network, as repeated high-frequency stimulus could trigger a continuous chain of LTP which infinitely increases the weights of the network (\cite{watt2010homeostatic}).  In order to attempt to counter this, we introduced both inhibitory neurons, as well as inhibitory plasticity, iSTDP, inspired by the implementation in \cite{litwin2014formation}.  The idea is that an inhibitory neuron should increase its firing to a post-synaptic excitatory neuron if that excitatory neuron fires too often.  

	As a consequence of STDP, certain "pathways" in the neural substrate begin to strengthen,; these are the cell assemblies that Hebb postulated.  Because they are more tightly connected, a lesser input is required to reignite the group.  In addition, because all the neurons are tightly bound, a reactivation in a small subset of the neurons will cause chain firing, ending up in the activation of the entire group\cite{holtmaat2016functional}.  A consequence of this is that if information can be successfully encoded in a neuronal group, it is extremely robust to noise and missing data.  
	
\begin{figure}
\includegraphics[scale=.50]{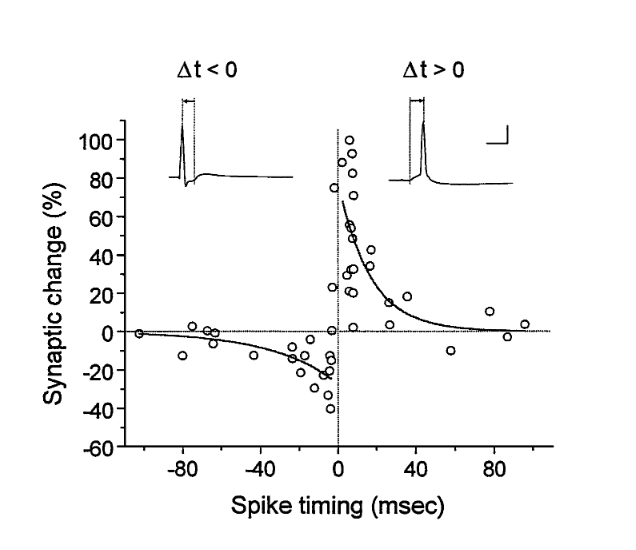}
\centering
\caption{A typical STDP curve.  Y axis is change in weight, X axis is time between post and pre spike. A negative time difference (post fires before pre) will decrease the weight, and the inverse for a positive one. (from Bi and Poo, 2001)}
\end{figure}

\section{Methods}

\subsection{DVS Camera}
The temporal input being investigated comes in the form of the biologically-inspired DVS camera (\cite{delbruck2008frame}).  The DVS sensor sends data in packets, called "events", rather than frames, and is a good candidate to take advantage of the independent nature of each neuron's input in an SNN.  The DVS camera is inspired by certain retinal ganglion cells that spike when a change in intensity is detected (\cite{sejnowski2012language}).  Similarly, rather than send all pixels at the same time, as with RGB cameras, the DVS camera sends only events that correspond to large changes in pixel intensity.  Because of the current instruction-based hardware, this network batches the events into discrete frames, although it could be transplanted into a neuromorphic chip hardware in the future that receives input asynchronously.  Recent work has been devoted to discovering good encoding mechanisms for DVS data for use in rate encoding network, whereas we make the argument that the raw DVS events themselves work well with the SNN network structure.

\begin{figure}
\includegraphics[scale=.70]{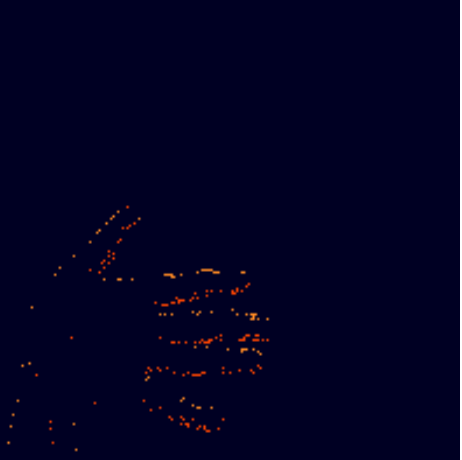}
\centering
\caption{Example DVS frame of a hand moving used in the input. The events capture the major shift in intensity of the edges of the hand as the hand moves across the background.}
\end{figure}


\subsection{Neuron Dynamics and Type}
The neuron activation is based off of the Izhikevich neuron (\cite{izhikevich2003simple}), which is a variant of the quadratic integrate and fire model; it is able to reproduce many behaviors seen by neurons in vivo, except is more computationally efficient than more complex models such as Hodgkin-Huxley (\cite{hodgkin1952currents}).  Specific neurotransmitter conductances are not modeled explicitly. At each time-step, the voltages for each neuron at each layer are updated according to the following equations:

\begin{equation}
v' = 0.04v^2 +5v + 140 - u + I_{syn} \\
\end{equation}
\begin{equation}
u' = a(bv - u)
\end{equation} 
	
with the voltage being reset to reset parameter $c$, and the variable $u$ being incremented by $d$, when the voltage goes over a threshold, $-30mV$, which indicates that the neuron spiked.  We used the same parameters as in \cite{izhikevich2003simple} to model a regular-spiking cortical neuron, with $a = .02$, $b = .2$, $c = -65mV$ and $d = 2$. In order to add randomness to the network, $c$ is modified by $15r^2$ and $d$ is modified by $-6r^2$, where r is a random number on the interval [0,1].  We did not model inhibitory neurons differently.  

At each time-step, every neuron $j$ updates its voltage as in eqn. 4, after calculating the incoming synaptic current.  For layer 1, this current is equal to 0 or 1, depending on if an event occurred from the DVS frame.  If the neuron is layers 2 or later, it calculates its voltage via the following:

\begin{equation}
I_{syn(i)} =  \sum_{j=0}^{k^2}w_{j,i}s_{j} + \sum_{l=0}^{m^2}w_{l,i}s_{l}
\end{equation}

where $I_{syn(i)}$ is the incoming synaptic current into neuron $i$, $k$ is the length of a side of the kernel for incoming connections and $m$ is the length for horizontal (see Figure. 3); $s_{x}$ is 1 if neuron $x$ spiked and 0 else.  Indices $j$ and $l$ are neurons that have synaptic connections to neuron $i$ from feed-forward and lateral connections, respectively. If a neuron is inhibitory, its weight is multiplied by $-1$. The neurons' weights are set within a bounds, [0, $w_{max}$] for excitatory and [$-w_{min}$,0] for inhibitory. 

\subsection{Network Structure}

The network consists of a set of connected "layers" of spiking neurons.  Each layer is a 2-dimensional sheet of neurons of the same $N x N$ dimensions (see Figure 3 for schematic). The code is completely custom-written, and is written expressly for the NVIDIA GPU using the CUDA programming language (\cite{luebke2008cuda}) to maximize speedup.  The advent of GPU technology allows large-scale networks like this to run in real time, even while training: a 5-layered network of 612,000 total neurons with 122 million total synapses, on the NVIDIA Titan Black, runs at 38.5 Hz without STDP updates, and 20Hz while training with STDP. 

The connections are inspired by how feed-forward connections in the mammalian brain are highly overlapping and convolutional (\cite{eickenberg2017seeing,lamme2000distinct}) and hierarchical (\cite{zeki1988functional}), being the inspiration for Convolutional Neural Networks (CNNs)\cite{lecun1998gradient}.  However, unlike CNN's, the kernels do not share weights, in an effort to remain biologically realistic, as well as to investigate if the non-shared quality of weights is an important factor in a spiking network.  The connection schema is loosely based on a mix of \cite{lumer1997neural} and modern CNN architectures.  Each neuron has feed-forward connections to a small square block of neurons in the post-layer, exactly as in a CNN.  This is done intentionally because the network is not learning a kernel using backpropagation, and needs as much variation as possible to aid in the STDP process.  In addition, in attempting to reverse-engineer the fundamental principles of spiking, it is biologically implausible that neurons communicate some sort of shared weight across the cortex; weight updates are local by nature. The weights are initialized with a uniform random distribution, and given an equally-high starting weight that allows spikes to occur so that STDP can begin reshaping the weights, consistent with theories that state that intrinsically high starting activation is a major part of neuronal group formation (\cite{holtmaat2016functional}). Excitatory weights are capped at +7 and inhibitory at -30, and STDP is only performed on feed-forward connections.
	In addition to the feed-forward connection,  each neuron also connects horizontally to a square of neurons in the same layer. This provides a degree of recurrence to the system, implementing a version of "re-entrant" connections as described by Edelman (\cite{edelman1987neural}), although more long-distance reentry is left for future work. We refer to the system as "semi-recurrent" and not fully recurrent because it does not contain backward connections to previous layers. 
	
	The network features both excitatory and inhibitory neurons, to maintain balance. As per prior work (\cite{izhikevich2004spike, litwin2014formation}) the ratio of excitatory to inhibitory is set at 4:1, to mirror the ratio of neurons seen in the mammalian brain. Only 20\% of feed-forward connections and 30\% of lateral connections were kept non-zero, per calculations done about the probability of connections in the cortex being between .1 and .3 (\cite{liley1994intracortical, braitenberg2013anatomy}).
	
\begin{figure}
\includegraphics[scale=.50]{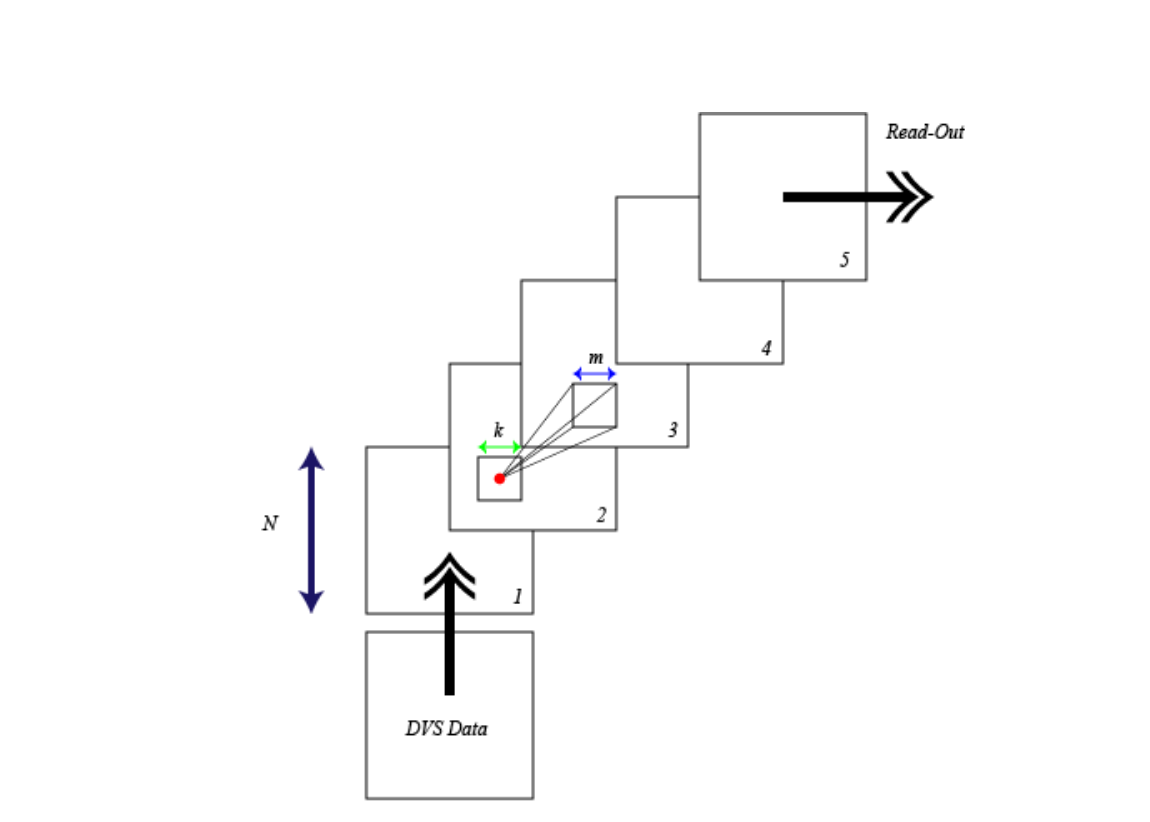}
\centering
\caption{The network architecture. Each layer is a 2-dimensional sheet of Izhikevich neurons of dimension $N$ x $N$. A given neuron, represented in red, has both horizontal connections to neurons in a $k$ x $k$ box around it in the same layer, as well as non-shared feed-forward connections in an $m$ x $m$ box in the next layer. The input DVS data is fed into the first layer, and the categorization is read out from groups formed in the last layer.  Layer Numbers are indicated in the bottom right corner.}
\end{figure}

\subsection{Network Information and Entropy}

Much work has been done to analyze the information capacity of neurons' firings using information-theoretic metrics. (\cite{borst1999information, butts2007temporal, van1997reproducibility}). Here, we use a metric, network entropy, that has been used in networks analyzing cancer genome alterations (\cite{teschendorff2010increased}), that analyzes the local entropy of the flux around each node.  To calculate the entropy for one neuron, we calculate an averaged entropy over all the outgoing connections, one entropy for inhibitory and one for excitatory: 

\begin{equation}
E_{i} = -1/log(K)\sum_{j \in N(i)}p_{i,j}log(p_{i,j}) 
\end{equation}

Where $E_{i}$ is the entropy for a given neuron $i$, $j$ is a neuron with a non-zero weight connected to neuron $i$, and $K$, the degree of $i$, is the number of non-zero connections to neuron $i$.  The mutual information given by an input is analogous to a decrease in entropy after a stimulus is presented(\cite{cover2012elements}).     
	
\section{Results and Discussion}

Fig. 4 shows some results after training the network continuously.  The top row shows the individual neurons spiking; the second row shows a spike count over a 300ms time window; the third and fourth row show the network entropy for each neuron.  When the network starts running, the entropy plot is completely black; all the weights are uniformly random, so the entropy is at maximum.  As STDP begins to strengthen the input patterns, and convolve itself forward into new patterns, it changes the weight distribution around a given neuron, thus altering the entropy. 

Due to the effects of the convolutional connections and STDP, rapid pulses of activity propagate through the network that do not occur without learning; this activity is similar to activity demonstrated in cortical-like networks in the brain (\cite{diesmann1999stable}). Because the weights are capped, and due to inhibitory signals and the voltage decay, this means that these neurons must be activated by concurrent, synchronous firing at precise times in order to cause the postsynaptic neuron to fire.  Unlike in \cite{litwin2014formation}, where uniformly low rates recruited depression, because the stimulus was driving the neurons, potentiation was dominant. The network exhibited local swells of activity in specific clusters, that would indicate a local synchrony, rather than a global synchrony (\cite{hosaka2008stdp}). Because of the lateral connections, these clusters were able to move horizontally across the same layer. The groups also seem to fire at higher rates, which would give evidence that they are in fact Hebbian groups. 

One point to note was that changing the parameters of the overlap size (the feed-forward convolution) had drastic effects on the activity of the network.  In addition, the maximum and minimum weight cutoffs were highly dependent on the size of the kernel: a larger kernel would overflow with activity with a much lower cutoff. 

One assumption is that the network entropy carries information about what the weights are learning.  The intution is that if a neuron were to have equal-weighted outgoing connections, it would offer no processing power, as all of the outputs would be the same and indistinguishable.   Because the network is learning a temporal signal, rather than a static one, we suppose that the network is learning "temporal features", in addition to traditional features such as lines, edges, and corners. Putting these together, we can see in Fig. 4 that as the spikes propagate through the network, they progress from being more local temporal-features, to more and more disperse; this would seem to correlate with the hypothesis that information is stored in a distributive fashion in the brain. These results would seem to aid credence to the hypothesis that temporally-precise firing can encode information about an input stimulus in a two-dimensional multi-layered network, because the synaptic weights encode the input stimulus. 

When inhibitory STDP is added, the entropy seems to be inverted, at least in the second row: where the entropy is high for excitatory neurons it is low for inhibitory ones.  We did not notice any difference in the firing of neural groups with the introduction of iSTDP, which is reflected in the relative similarity of the excitatory entropy between figures 4 and 5; this could either be because of the weight cutoff, or because inhibition works less in later layers. For both observations, they will be an interesting avenue of investigation for future work.

\subsection{Future Directions}

A realistic first step would be to perform simple machine learning techniques on the last layer of the network, such as using a linear SVM, to see if the features encoded in the last layer can predict the class of input.  Knowing that the network organizes its weights around the input stimulus, the next step is to integrate these networks into larger networks that could use the output of the final layers for further processing.  For example, the last layer could be connected to a 1-D layer of lateral-inhibition neurons, which would act as a winner-take-all classification layer.  Such neurons could be linked up to motor commands, for example moving a hand, and the network could be used to explore and validate any role STDP has in motor learning in development, both as investigation into humans as well as to train dexterous robots for quick and smooth maneuvers.     

\begin{figure}
\includegraphics[scale=.55]{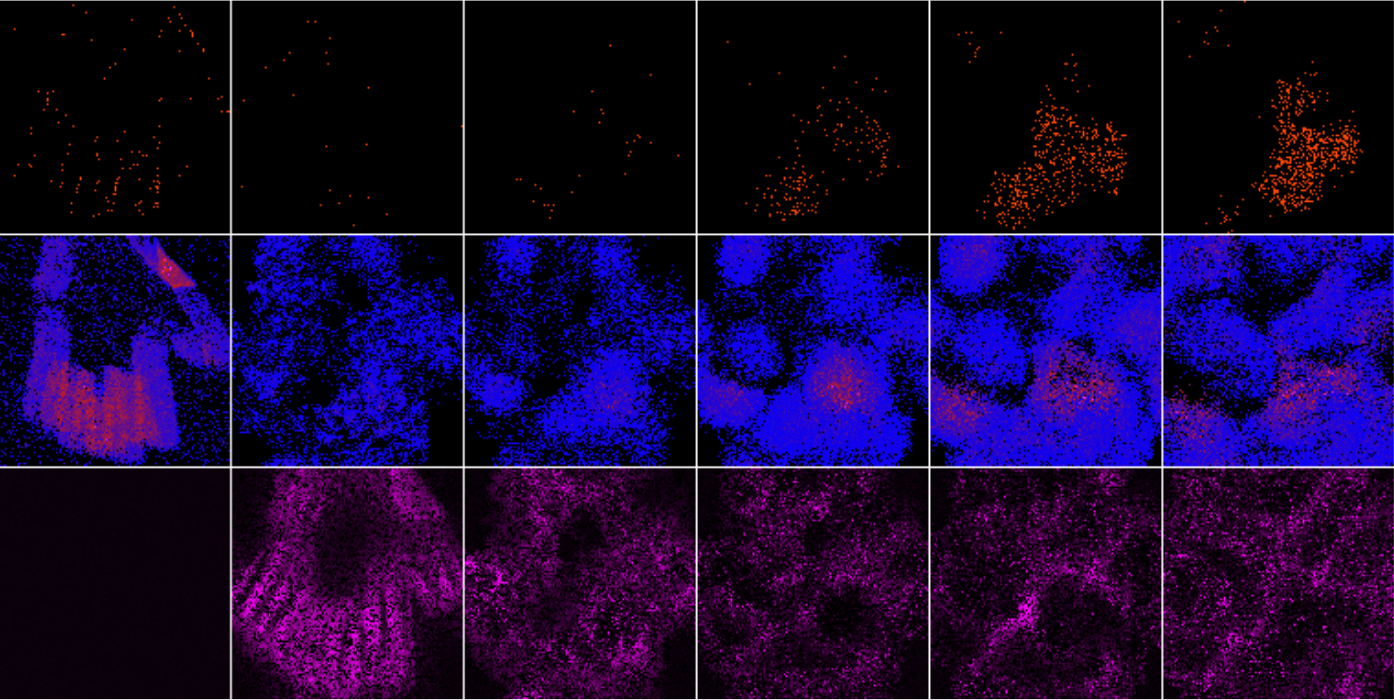}
\centering
\caption{A display of the network running. The network consists of 96,774 neurons and 7.5 million synapses. Here we have turned off iSTDP; just excitatory STDP is running. The top row displays the firing, the second an average spike count analagous to a PSTH that gets refreshed every 300ms (red = high rate, blue = low), and the last row displays the network entropy, darker colors represent higher entropy for a given neuron. Also visible is a feed-forward propagation of activity from lower layers.}
\end{figure}

\begin{figure}
\includegraphics[scale=.55]{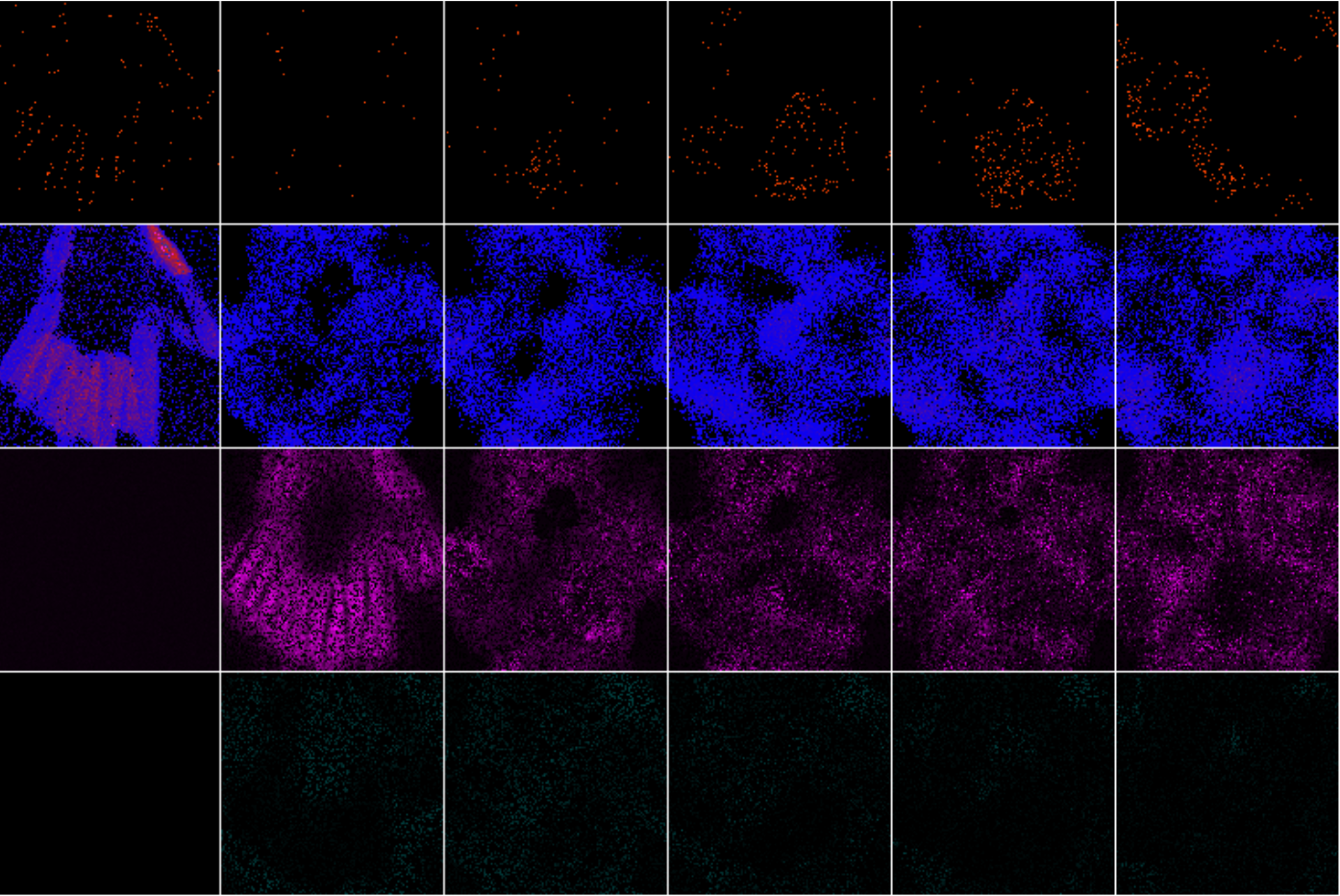}
\centering
\caption{A display of the network running with iSTDP, inhibitory entropy is in the bottom row.}
\end{figure}

\section{Conclusion}
Here we have demonstrated a novel semi-recurrent feed-forward dynamical network and provided evidence that it is encoding information about temporal stimuli.  Our hope is that this will lead to future work that takes advantage of the potential of temporal coding in spiking networks.  

\subsubsection*{Acknowledgments}

This work was supported partly by the University of Maryland COMBINE program and NSF award DGE-1632976. The authors would like to also thank Mrs. Greg Davis and Jesse Milzman at the University of Maryland for thoughtful conversation and advice.

\clearpage

\bibliography{NIPS_COMPNEURO_EVANUSA.bib}

\end{document}